\documentclass[journal,transmag]{IEEEtran}
\usepackage{amsmath,amssymb,graphicx}
\usepackage[caption=false,font=footnotesize]{subfig}
\usepackage{hyperref}

\usepackage[thmmarks, amsmath]{ntheorem}
\makeatletter \renewtheoremstyle{plain} {\item{\theorem@headerfont ##1\ ##2\theorem@separator}~}  {\item{\theorem@headerfont ##1\ ##2\ (##3)\theorem@separator}~}
\theoremheaderfont{\normalfont\bfseries}
\theoremseparator{:}
\theorembodyfont{\normalfont}
\theoremsymbol{\ensuremath{\blacksquare}}
\newtheorem{definition}{Definition}

\usepackage{algorithm}
\usepackage{algpseudocode}

\date{}
\title{PCM and APCM Revisited: An Uncertainty Perspective}

\begin{document}

\author{\IEEEauthorblockN{Peixin~Hou\IEEEauthorrefmark{1}, Hao~Deng\IEEEauthorrefmark{2}, Jiguang~Yue\IEEEauthorrefmark{1}, and Shuguang~Liu\IEEEauthorrefmark{3}}
\IEEEauthorblockA{\IEEEauthorrefmark{1}Department of Control Science and Engineering, Tongji University}
\IEEEauthorblockA{\IEEEauthorrefmark{2}School of Physics and Electronics, Henan University}
\IEEEauthorblockA{\IEEEauthorrefmark{3}Department of Hydraulic Engineering, Tongji University}%
}

\IEEEtitleabstractindextext{%
\begin{abstract}
In this paper, we take a new look at the possibilistic c-means (PCM) and adaptive PCM (APCM) clustering algorithms from the perspective of uncertainty. This new perspective offers us insights into the clustering process, and also provides us greater degree of flexibility. We analyze the clustering behavior of PCM-based algorithms and introduce parameters $\sigma_v$ and $\alpha$ to characterize uncertainty of estimated bandwidth and noise level of the dataset respectively. Then uncertainty (fuzziness) of membership values caused by uncertainty of the estimated bandwidth parameter is modeled by a conditional fuzzy set, which is a new formulation of the type-2 fuzzy set. Experiments show that parameters $\sigma_v$ and $\alpha$ make the clustering process more easy to control, and main features of PCM and APCM are unified in this new clustering framework (UPCM). More specifically, UPCM reduces to PCM when we set a small $\alpha$ or a large $\sigma_v$, and UPCM reduces to APCM when clusters are confined in their physical clusters and possible cluster elimination are ensured. Finally we present further researches of this paper.
\end{abstract}

}

\maketitle

\section{Introduction}
\label{sec-1}
The goal of clustering analysis is to discover natural groups of a set of points according to intrinsic characteristics or similarity \cite{jain_data_2010}. A large portion of clustering algorithms are prototype-based, i.e., clusters are represented by vectors in the data space. Prototype-based clustering algorithms aim to move prototypes into dense data regions. So we can say that the prototype (cluster center) represents the cluster.

However, the performance of clustering algorithms can be severely affected by uncertainty in the dataset. Generally, uncertainty exits in two forms: noisy points and the parameters estimated from noisy points.

The degree of confidence in data are treated differently in clustering algorithms.
K-means \cite{jain_data_2010} assumes that all data are reliable. It is sensitive to outliers and noise. Even if an object is quite far away from the cluster center, it is still forced into a cluster, which distorts the cluster shapes. ISODATA \cite{ball_clustering_1967} and PAM \cite{kaufman_finding_2009} both consider the effect of outliers in clustering procedures. ISODATA gets rid of clusters with few objects. PAM utilizes real data points (medoids) as the cluster prototypes and avoids the effect of outliers \cite{xu_survey_2005}.
Fuzzy c-means (FCM) \cite{bezdek_pattern_2013} is also sensitive to noise and outliers, and the resulted memberships are relative values.  Possibilistic c-means (PCM) \cite{krishnapuram_possibilistic_1993} forces the memberships to decrease with the distance from cluster center. It's shown in \cite{dave_robust_1997} that PCM can be also interpreted as keeping $c\text{-noise}$ clusters to ensure that the memberships of outliers are reduced. Furthermore, the memberships are controlled by a bandwidth parameter estimated from data.

On the other hand, the uncertainty of parameters estimated from noisy data is rather complicated to handle.
Adaptive PCM (APCM) \cite{xenaki_novel_2016} considers the estimated bandwidth to be unreliable and introduces a hyper-parameter specified by the user to manually correct the bandwidth. In fact, the uncertainty of bandwidth can be handled in a more fuzzy way because the uncertainty of memberships is caused by the uncertainty of the estimated bandwidth.
As we know, the type-2 fuzzy set \cite{zadeh_concept_1975}\cite{mendel_type-2_2002} allows us to model one fuzziness over another fuzziness. Recently, the conditional fuzzy set framework \cite{wang_new_2016} of a type-2 fuzzy set is proposed so that we can explicitly model the uncertainty of memberships caused by some parameter.

In this paper, we take a new look at PCM and APCM clustering algorithms from the perspective of uncertainty. This new perspective offers us insights into the clustering process, and also provides us greater degree of flexibility.
Our contributions are summarized as follows:
\begin{enumerate}
\item The uncertainty (fuzziness) of the estimated bandwidth parameter is specified by a hyper-parameter $\sigma_v$, and the uncertainty (fuzziness) of membership values caused by the estimated bandwidth parameter is modeled by a conditional fuzzy set.
\item As to PCM-based clustering algorithms, we propose that each cluster can be seen as noise to other clusters. This conclusion comes from the fact that the existence of close physical clusters makes clustering difficult. This new form of noise is characterized by noise level $\alpha$ of the data set.
\item The introduction of parameters $\sigma_v$ and $\alpha$ reformulates PCM and APCM from the perspective of uncertainty. Furthermore, main features of PCM and APCM are unified in this new clustering framework.
\item Experiments show that the bandwidth uncertainty $\sigma_v$ depends on noise level $\alpha$.
This fact reveals that uncertainty of the estimated bandwidth can also be modeled by a type-2 fuzzy set and its uncertainty is caused by $\alpha$. The dependence of $\sigma_v$ on $\alpha$ can be studied in further researches.
\end{enumerate}

The rest of this paper is organized as follows. Section \ref{sec-2} analyzes the clustering behavior of PCM-based algorithms and presents the motivations of this paper. Section \ref{sec-3} introduces how to incorporate uncertainty of the estimated bandwidth into uncertainty of the membership function. In Section \ref{sec-4}, we present the unified framework (UPCM) of PCM and APCM. In Section \ref{sec-5}, we discuss the difference between APCM and UPCM in handling uncertainty. Further researches are also presented.
\section{Review of PCM and APCM, and Motivations}
\label{sec-2}
In this section, we first review the possibilistic c-means (PCM) and adaptive PCM (APCM) clustering algorithms. Then we analyze two typical clustering problems and present the motivations of this paper.
\subsection{PCM and APCM Review}
\label{sec-2-1}
Typicality is one of the most commonly used interpretations of memberships in applications of fuzzy set theory. The membership value produced by fuzzy c-means (FCM) \cite{bezdek_pattern_2013} can't be used to indicate the typicality of a point in the cluster. Possibilistic c-means (PCM) \cite{krishnapuram_possibilistic_1993} solves this problem by forcing the membership of a point to be small if it's far from the cluster center (prototype). This intuition is incorporated into the objection function by adding a penalty term:
\begin{equation}
J(\mathbf{\Theta},\mathbf{U})=\sum_{j=1}^{c}J_j=\sum_{j=1}^{c}\left[\sum_{i=1}^{N}u_{ij}d_{ij}^2+\gamma_j \sum_{i=1}^{N}f(u_{ij})\right]
\end{equation}
where $\mathbf{\Theta}=(\boldsymbol{\theta}_1,\ldots,\boldsymbol{\theta}_c)$ is a $c$-tuple of prototypes, $d_{ij}$ is the distance of feature point $\mathbf{x}_i$ to prototype $\boldsymbol{\theta}_j$, $N$ is the total number of feature vectors, $c$ is the number of clusters, and $\mathbf{U}=[u_{ij}]$ is a $N\times c$ matrix where $u_{ij}$ denotes the \emph{degree of compatibility} of $\mathbf{x}_i$ to the $j\text{th}$ cluster $C_j$ which is represented by $\boldsymbol{\theta}_j$. $\gamma_j$ can be seen as a bandwidth parameter of the possibility (membership) distribution for each cluster. Note that either $\gamma_j$ or $\sqrt{\gamma_j}$ can be referred to as the bandwidth for convenience in this paper. $f(\cdot)$ is a decreasing function of $u_{ij}$ and forces the $u_{ij}$ as large as possible, thus avoiding the trivial solution that $u_{ij}=0$. A good choice for $f(\cdot)$ is proposed in \cite{krishnapuram_possibilistic_1996}:
\begin{equation}
f(u_{ij})=u_{ij}\log u_{ij}-u_{ij}
\end{equation}

After minimizing $J(\mathbf{\Theta},\mathbf{U})$ with respect to $u_{ij}$ and $\boldsymbol{\theta}_j$, we get the following update equations:
\begin{IEEEeqnarray}{ll}
u_{ij}&=\exp\left(-\frac{d^2_{ij}}{\gamma_j}\right) \label{pcm_u_update}  \\
\boldsymbol{\theta}_j&=\frac{\Sigma_{i=1}^Nu_{ij}\mathbf{x}_i}{\Sigma_{i=1}^Nu_{ij}} \label{pcm_theta_update}
\end{IEEEeqnarray}

In PCM, clusters do not have a lot of mobility, so a reasonable good initialization is required for the algorithm to converge to the global minimum. A common strategy for initializing is to run the FCM algorithm first and set
\begin{equation}
\gamma_j=\frac{\Sigma_{i=1}^Nu_{ij}^{FCM}d^2_{ij}}{\Sigma_{i=1}^Nu_{ij}^{FCM}}
\end{equation}
where $d_{ij}=||\mathbf{x}_i-\boldsymbol{\theta}_j||$, $\boldsymbol{\theta}_j\text{s}$ and $u_{ij}^{FCM}\text{s}$ are the final FCM estimates for cluster prototypes and membership values respectively. Then $\gamma_j\text{s}$ are fixed and iterations are performed until a specific termination criterion is met.

As pointed out in \cite{krishnapuram_possibilistic_1996}, PCM is primarily a mode-seeking algorithm. In other words, the algorithm can potentially find $c$ dense regions from a data set that may not have $c$ clusters. However, due to the well-known fact that since no link exists among clusters, each $J_j$ can be minimized independently, the $c$ dense regions found may be coincident, as reported in \cite{barni_comments_1996}. It was suggested in \cite{krishnapuram_possibilistic_1996} that this behavior is "a blessing in disguise" and can be utilized by merging coincident clusters after over-specifying $c$. This idea is implemented in the adaptive possibilistic c-means algorithm (APCM) \cite{xenaki_novel_2016} by adapting $\gamma_j$ at each iteration. Cluster $C_j$ is merged with another cluster and is eliminated when there are no points in cluster $C_j$ or $\gamma_j$ becomes $0$. This cluster elimination ability allows us to over-specify the cluster number and the algorithm still produces a reasonable number of clusters, which makes the algorithm very flexible because we don't need to have strong prior knowledge of the cluster number.

However we should prevent the unexpected cluster elimination. In the case where two physical clusters with very different variance are located very close to each other (see Fig.\ref{fig1_ori}), the prototype of the small variance cluster is affected by the data points of its nearby big cluster which has numerous points, according to \eqref{pcm_u_update} and \eqref{pcm_theta_update}. As a result, the two prototypes will merge. PCM can't handle this problem because it has no corresponding parameters to control the clustering process. APCM alleviates this issue by introducing a parameter in $\gamma_j$ to manually scale the bandwidth:
\begin{equation}
\label{corrected_eta}
\gamma_j=\frac{\hat{\eta}}{\alpha}\eta_j
\end{equation}
where $\hat{\eta}$ is a constant defined as the minimum among all initial $\eta_j\text{s}$, $\hat{\eta}=\min_j\eta_j$, and $\alpha$ is chosen so that the quantity $\hat{\eta}/\alpha$ equals to the mean absolute deviation ($\eta_j$)  of the smallest physical cluster formed in the dataset. $\eta_j$ is initialized as
\begin{equation}
\label{apcm_eta_init}
\eta_j=\frac{\Sigma_{i=1}^Nu_{ij}^{FCM}d_{ij}}{\Sigma_{i=1}^Nu_{ij}^{FCM}}
\end{equation}
where $d_{ij}=||\mathbf{x}_i-\boldsymbol{\theta}_j||$, $\boldsymbol{\theta}_j\text{s}$ and $u_{ij}^{FCM}\text{s}$ in \eqref{apcm_eta_init} are the final parameter estimates obtained by FCM. $\eta_j$ is updated at each iteration as the \emph{mean absolute deviation} of the most compatible to cluster $C_j$ data points which form a set $A_j$, i.e., $A_j=\{\mathbf{x}_i|u_{ij}=\max_r u_{ir}\}$.
\begin{equation}
\label{apcm_eta_update}
\eta_j=\frac{1}{n_j}\sum_{\mathbf{x}_i\in A_j}||\mathbf{x}_i-\boldsymbol{\mu}_j||
\end{equation}
where $n_j$ and $\boldsymbol{\mu}_j$ are the number of points in $A_j$ and the mean vector of points in $A_j$ respectively. APCM only allows points in $A_j$ to update $\eta_j$, which is an essential condition for succeeding cluster elimination, as by this way, $\eta_j$ can decrease to $0$. APCM chooses $\boldsymbol{\mu}_j$ instead of $\boldsymbol{\theta}_j$ to update $\eta_j$ because $\boldsymbol{\theta}_j$ may vary significantly while $\boldsymbol{\mu}_j$ is more stable during the first few iterations.
\subsection{Motivations}
\label{sec-2-2}
\begin{figure}[!t]
   \centering
   \subfloat[]
    {\includegraphics[width=0.5\columnwidth]{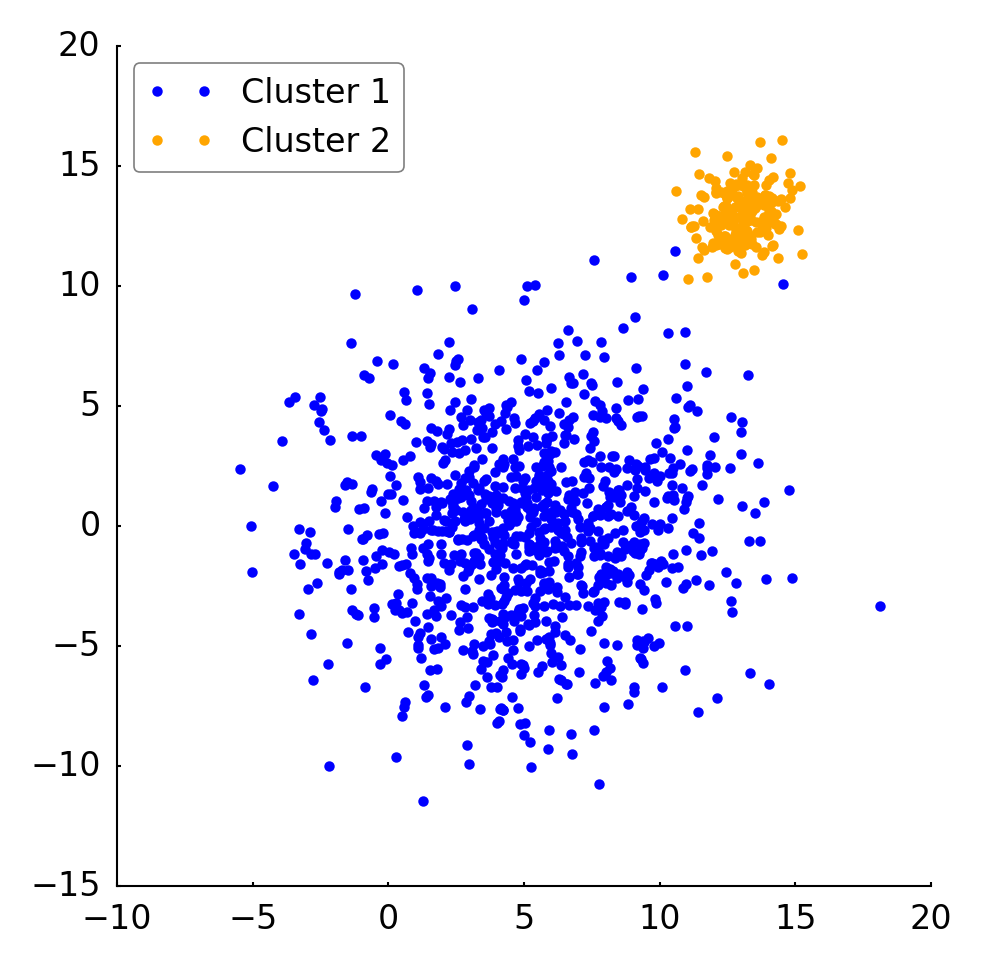}\label{fig1_ori}}
   \subfloat[]
    {\includegraphics[width=0.5\columnwidth]{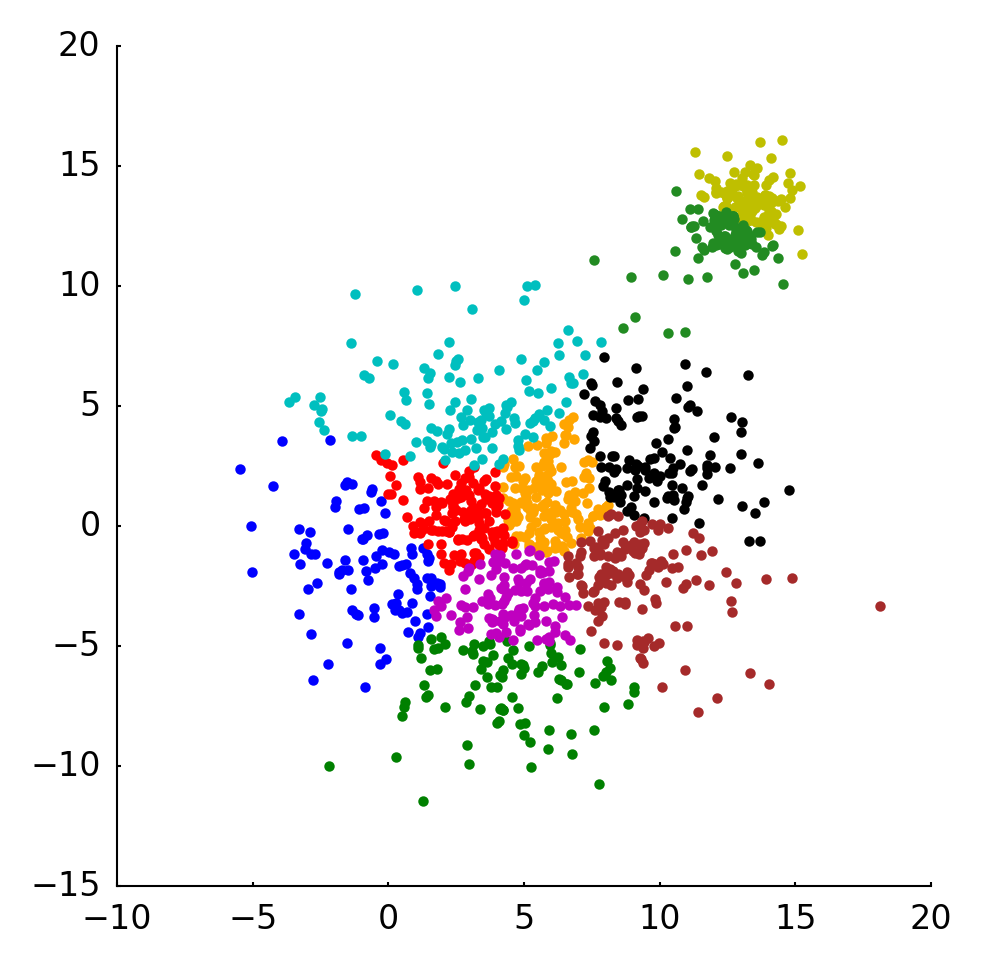}\label{fig1_init}}
\caption{(a) Dataset 1. (b) 10 initial partitions obtained by FCM.}
\label{fig1}
\end{figure}
\begin{figure}[!t]
   \centering
   \subfloat[]
    {\includegraphics[width=0.5\columnwidth]{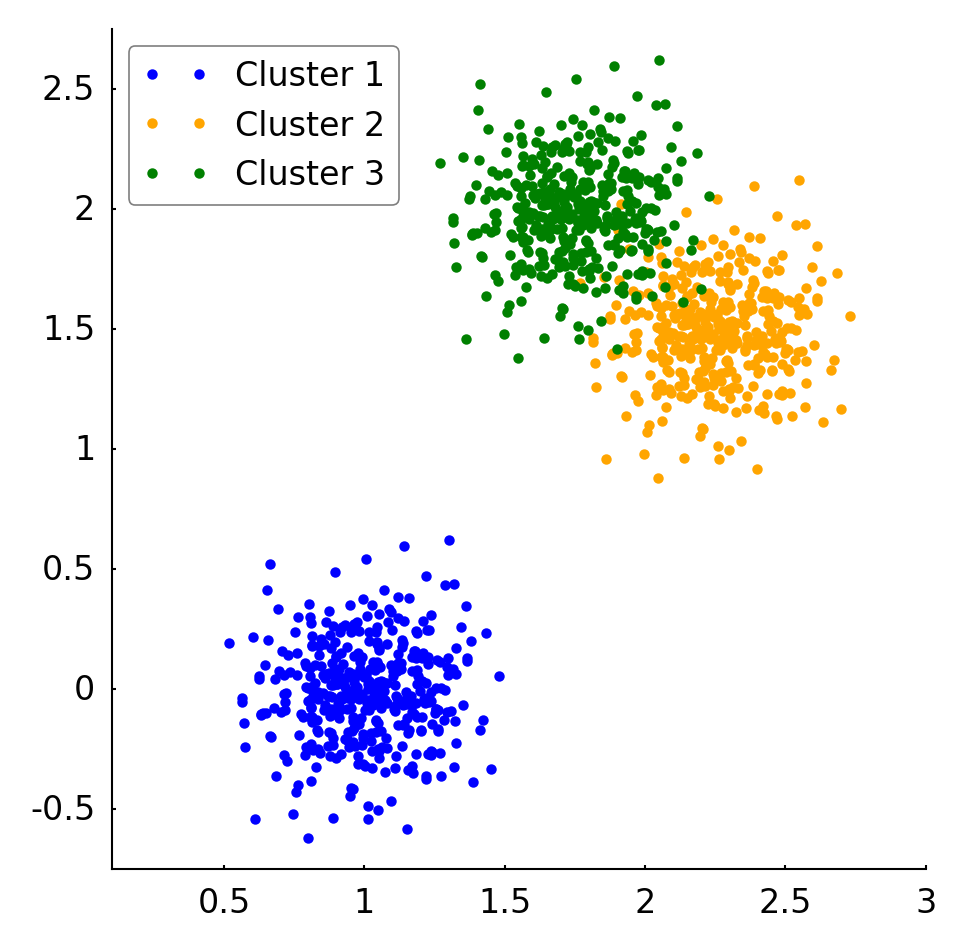}\label{fig6_ori}}
   \subfloat[]
    {\includegraphics[width=0.5\columnwidth]{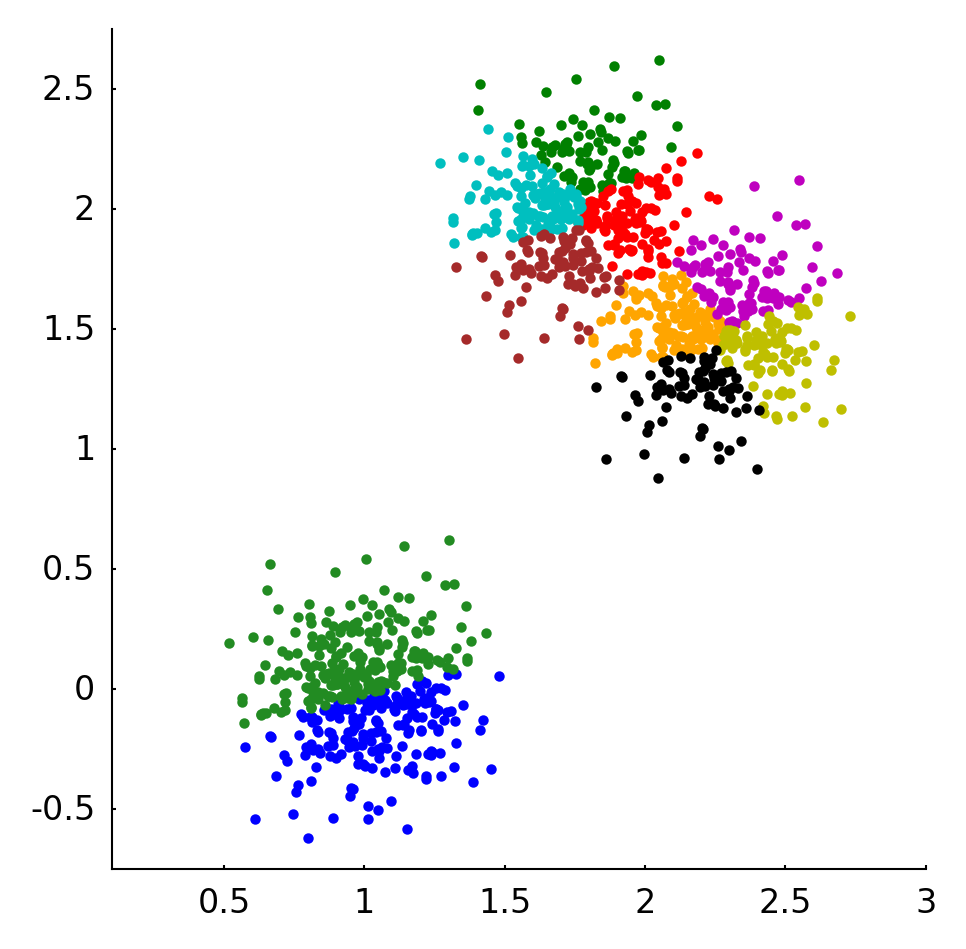}\label{fig6_init}}
\caption{(a) Dataset 2. (b) 10 initial partitions obtained by FCM.}
\label{fig6}
\end{figure}
Fig.\ref{fig1_ori} and Fig.\ref{fig6_ori} are two typical clustering problems. The two clusters in Fig.\ref{fig1_ori} are generated by normal distributions with centers $\mathbf{c_1}=[13, 13]^T$, $\mathbf{c_2}=[5, 0]^T$, covariance matrixes $\mathbf{\Sigma_1}=\mathbf{I}$, $\mathbf{\Sigma_2}=3.7^2\mathbf{I}$, $N_1=200$ points, and $N_2=1000$ points  respectively, where $\mathbf{I}$ is the $2\times 2$ identity matrix. The three clusters in Fig.\ref{fig6_ori} are generated by normal distributions with  centers $\mathbf{c_1}=[1, 0]^T$, $\mathbf{c_2}=[2.25, 1.5]^T$, $\mathbf{c_3}=[1.75, 2]^T$ respectively, all with $N=400$ points, and covariance matrixes are all $\mathbf{\Sigma}=0.2^2\mathbf{I}$. Fig.\ref{fig1_init} and Fig.\ref{fig6_init} show the initialization results obtained by FCM with 10 clusters. Note that the initial clusters in each physical cluster should merge, while clusters in different physical clusters shouldn't.

The two physical clusters in Fig.\ref{fig1} are well separated. With the initialization of Fig.\ref{fig1_init}, APCM estimates $\eta_j$ via \eqref{apcm_eta_update}, which is corrected by $\hat{\eta}/\alpha$ and we get the bandwidth $\gamma_j$ via \eqref{corrected_eta}. The only care is that the bandwidth correction term $\hat{\eta}/\alpha$ specified by the user is not too small so that the small initialization clusters of Cluster $1$ have enough mobility to move to the dense region of each physical cluster and  merge, according to \eqref{pcm_theta_update} (Note that large bandwidth means more mobility and to avoid the case where all clusters merge into one cluster, $\hat{\eta}/\alpha$ also should not be too large. This fact can be seen in Fig.7 of \cite{xenaki_novel_2016} when $\alpha$ is small).
As to Fig.\ref{fig6}, Cluster $2$ and Cluster $3$ are not well separated, so we should take more care. The bandwidth correction term $\hat{\eta}/\alpha$ should not be too small so that the small initialization clusters of each physical cluster can merge. The term $\hat{\eta}/\alpha$ also shouldn't be too large so that Cluster $2$ and Cluster $3$ don't have enough mobility to merge.
In summary, the choice of $\alpha$ in the correction term should be dealt with differently. And two needs naturally arise from the above observation.
\begin{enumerate}
\item We should control the clustering process basing on the noise level of the dataset. The above analysis shows that there is some difference between the two problems. In fact, the clustering algorithm faces a more noisy environment in Fig.\ref{fig6_ori} than in Fig.\ref{fig1_ori} in the sense that there are close clusters in dataset Fig.\ref{fig6_ori}.
\item We should have a more flexible bandwidth correction technique.
The reason APCM introduces a bandwidth correction term is that the estimated bandwidth is not always reliable to recognize the structure underlying the data set.
In other words, the estimated bandwidth is uncertain, and this uncertainty causes the membership value of a point to be uncertain (see \eqref{pcm_u_update}), then the cluster center also becomes uncertain (see \eqref{pcm_theta_update}). If this uncertainty is not properly handled, the clustering algorithm would fail.
In fact, the bandwidth estimation uncertainty can be attributed to the noise in data points.
In APCM, membership values of all points in each cluster are treated equally uncertain, and receive the same bandwidth correction.
However, the uncertainties are different, i.e., we have more confidence in the estimated membership value of a point if this point is near the prototype (cluster center).
So we should correct the bandwidth in a more reasonable way.
\end{enumerate}

This paper aims to address the above two needs.
To address the second need, Section \ref{sec-3} shows how to use the type-2 fuzzy set to incorporate uncertainty of the estimated bandwidth into the membership value of point $\mathbf{x}_i$. As will be seen in Section \ref{sec-4}, the first need is addressed by introducing a noise level parameter so that Fig.\ref{fig1} and Fig.\ref{fig6} can be treated differently. After addressing these two needs, we reformulate PCM and APCM into the same framework (UPCM).
\section{The Conditional Fuzzy Set Framework}
\label{sec-3}
In this section, we first review the conditional fuzzy set framework. Then we show through an example that this new definition of a type-2 fuzzy is natural and reasonable to incorporate the uncertainty of the estimated bandwidth.
\subsection{The Conditional Fuzzy Set Framework Review}
\label{sec-3-1}
According to Zadeh \cite{zadeh_concept_1975}, a type-2 fuzzy set (T2 FS) is a fuzzy set whose membership values are type-1 fuzzy sets on $[0,1]$. When written in more precise mathematical terms, this definition becomes as follows \cite{wang_new_2016}:

\begin{definition}[type-2 fuzzy sets]
\label{type2-fs}
A type-2 fuzzy set $\tilde{X}$ is a fuzzy set defined on the universe of discourse $\Omega_X$ whose membership value $\mu_{\tilde{X}}(x)$ for a given $x\in\Omega_X$ is a type-1 fuzzy set  $U(x)=\mu_{\tilde{X}}(x)$ defined on $\Omega_X\subseteq[0,1]$ with membership function $\mu_{U(x)}(x,\mu_x)$ where $\mu_x\in\Omega_X\subseteq[0,1]$. The x is called \emph{primary variable} and $\mu_x$ is called the \emph{secondary variable}.
\end{definition}

It's clear that T2 FS is just that one fuzziness (uncertainty) depends on another fuzziness. However Definition \ref{type2-fs} makes T2 FS a complex subject. To simplify this problem, Li-Xin Wang \cite{wang_new_2016} proposes a conditional fuzzy set framework:

\begin{definition}[conditional fuzzy sets]
\label{conditional-fs}
Let $X$ and $V$ be fuzzy sets defined on $\Omega_X$ and $\Omega_Y$, respectively. A \emph{conditional fuzzy set}, denoted as $X|V$, is a fuzzy set defined on $\Omega_X$ with membership function:
\begin{equation}
\mu_{X|V}(x|V),\quad  x\in\Omega_X
\end{equation}
depending on the fuzzy set $V$ whose membership function is $\mu_V(v)$ with $v\in\Omega_V$. The x is called the \emph{primary variable} and $v$ is called the \emph{secondary variable}; the membership function $\mu_{X|V}(x|V)$ characterizes the \emph{primary fuzziness} while the membership function $\mu_V(v)$ characterizes the \emph{secondary fuzziness}.
\end{definition}

This framework resembles the concept of conditional probability in probability theory, which studies the dependence of one randomness on the other randomness. It is shown in \cite{wang_new_2016} that the above two definitions are equivalent. However the conditional fuzzy set framework provides a much more natural framework to model the dependence among multiple fuzziness than the type-2 fuzzy set formulation.
In most real-world applications we choose the membership functions to have a fixed structure with some free parameters, such as the Gaussian membership function with the center or standard deviation as free parameters. In such formulations, the uncertainty (fuzziness) of the membership comes from the uncertainties of the free parameters; i.e., the parameter uncertainties are the causes, while the membership uncertainty is the effect, and it is natural to choose the independent cause as the secondary variable to characterize the secondary fuzziness (as in Definition \ref{conditional-fs} for a conditional fuzzy set), rather than choosing the dependent effect as the secondary variable (as in Definition \ref{type2-fs} for a type-2 fuzzy set).

It is also shown in \cite{wang_new_2016} that a conditional fuzzy set $X|V$ is equivalent to a fuzzy relation \cite{wang_course_1997} on $\Omega_X\times\Omega_V$ with membership function:
\begin{equation}
\label{fuzzy_relation}
\mu_{X|V}(x,v)=t[\mu_{X|V}(x|v),\mu_V(v)]
\end{equation}
where $x\in\Omega_X$, $v\in\Omega_V$, $t[*,*]$ is the $t$-norm operator with minimum and product as the most common choices, and $\mu_{X|V}(x,v)$ is the membership function $\mu_{X|V}(x|V)$ of the conditional fuzzy set $X|V$ with the fuzzy set $V$  replaced by a free variable $v\in\Omega_V$.

In the study of several random variables, the statistics of each are called marginal, and the probability density function (pdf) of a single random variable is called a marginal pdf. Similarly, since the conditional fuzzy set or the type-2 fuzzy set contains two fuzzy variables (the primary and secondary variables), the concept of marginal fuzzy set for conditional fuzzy sets is introduced in \cite{wang_new_2016} as follows:

\begin{definition}[marginal fuzzy sets, Compositional Rule of Inference Scheme]
\label{marginal-fs}
Let $X|V$ be a conditional fuzzy set defined in Definition \ref{conditional-fs} whose membership function $\mu_{X|V}(x,v)$ is given by \eqref{fuzzy_relation}. The \emph{marginal fuzzy set} of $X|V$, denoted as $X$, is a type-1 fuzzy set on $\Omega_X$ whose membership function $\mu_X(x)$ is determined through Zadeh's Compositional Rule of Inference:
\begin{equation}
\label{marginal_fs}
\mu_X(x)=\max_{v\in\Omega_V}\min[\mu_{X|V}(x|v),\mu_V(v)],\;\;x\in\Omega_X.
\end{equation}
\end{definition}

Then the basic philosophy to deal with type-2 fuzziness is to use \eqref{marginal_fs} to "cancel out" the secondary fuzziness $V$ and transform the type-2 problems back to the ordinary type-1 framework. We can explicitly model the uncertainty of the membership caused by some parameter $V$ and "cancel" $V$ to get the type-1 marginal fuzzy set. Then the effect of the uncertainty of $V$ is incorporated into type-1 marginal fuzzy set.
\subsection{An Example to Illustrate the Incorporation of Uncertainty}
\label{sec-3-2}
Suppose we have estimated the one-dimensional center $x_0$ and bandwidth $v_0$ of a Gaussian membership function $\mu_X(x)$ to represent some cluster, and we want to consider the uncertainty of $\mu_X(x)$ caused by the uncertainty of the bandwidth parameter $V$. First, the conditional fuzzy set $X|V$ is constructed as follows:
\begin{equation}
\mu_{X|V}(x|V)=\exp\left(-\frac{|x-x_0|^2}{V^2}\right)
\end{equation}
and the uncertainty (fuzziness) of $V$ is also modeled as a Gaussian fuzzy set with the membership function:
\begin{equation}
\label{secondary_fuzziness_v}
\mu_V(v)=\exp\left(-\frac{(v-v_0)^2}{\sigma^2_v}\right)
\end{equation}
where $\sigma_v$ is a given constant which represents the uncertainty of parameter $V$. Then according to Definition \ref{marginal-fs} \eqref{marginal_fs}, the marginal fuzzy set $X$ of $X|V$ with membership function:
\begin{IEEEeqnarray}{ll}
\label{marginal_result}
\mu_X(x)&=\max_{v\in R_+ }\min\left[\exp\left(-\frac{|x-x_0|^2}{v^2}\right),\exp\left(-\frac{(v-v_0)^2}{\sigma^2_v}\right)\right] \nonumber \\
        &=\exp\left(-\frac{|x-x_0|^2}{v_{new}}\right)
\end{IEEEeqnarray}
where $v_{new}=\left(0.5v_0+0.5\sqrt{v_0^2+4\sigma_v|x-x_0|}\right)^2$.
The last step is achieved at the highest point of the intersection $\exp\left(-\frac{|x-x_0|^2}{v^2}\right)=\exp\left(-\frac{(v-v_0)^2}{\sigma^2_v}\right)$ which gives:
\begin{IEEEeqnarray*}{ll}
v_{new1} &= 0.5v_0+0.5\sqrt{v_0^2+4\sigma_v|x-x_0|}\geq v_0, \\
v_{new2} &= 0.5v_0-0.5\sqrt{v_0^2-4\sigma_v|x-x_0|}\leq v_0.
\end{IEEEeqnarray*}
Then we get \eqref{marginal_result} through
\begin{equation*}
\max\left[\exp\left(-\frac{|x-x_0|^2}{v_{new1}^2}\right),\exp\left(-\frac{|x-x_0|^2}{v_{new2}^2}\right)\right]=\exp\left(-\frac{|x-x_0|^2}{v_{new1}^2}\right).
\end{equation*}
Let $d(\mathbf{x}_i,\mathbf{x}_0)$ denote the distance from a point $\mathbf{x}_i$ to the center $\mathbf{x}_0$. Then result \eqref{marginal_result} can be generalized by replacing $|x-c|$ with $d(\mathbf{x}_i,\mathbf{x}_0)$.
\begin{figure}[!t]
   \centering
   \subfloat[]
    {\includegraphics[width=0.5\columnwidth]{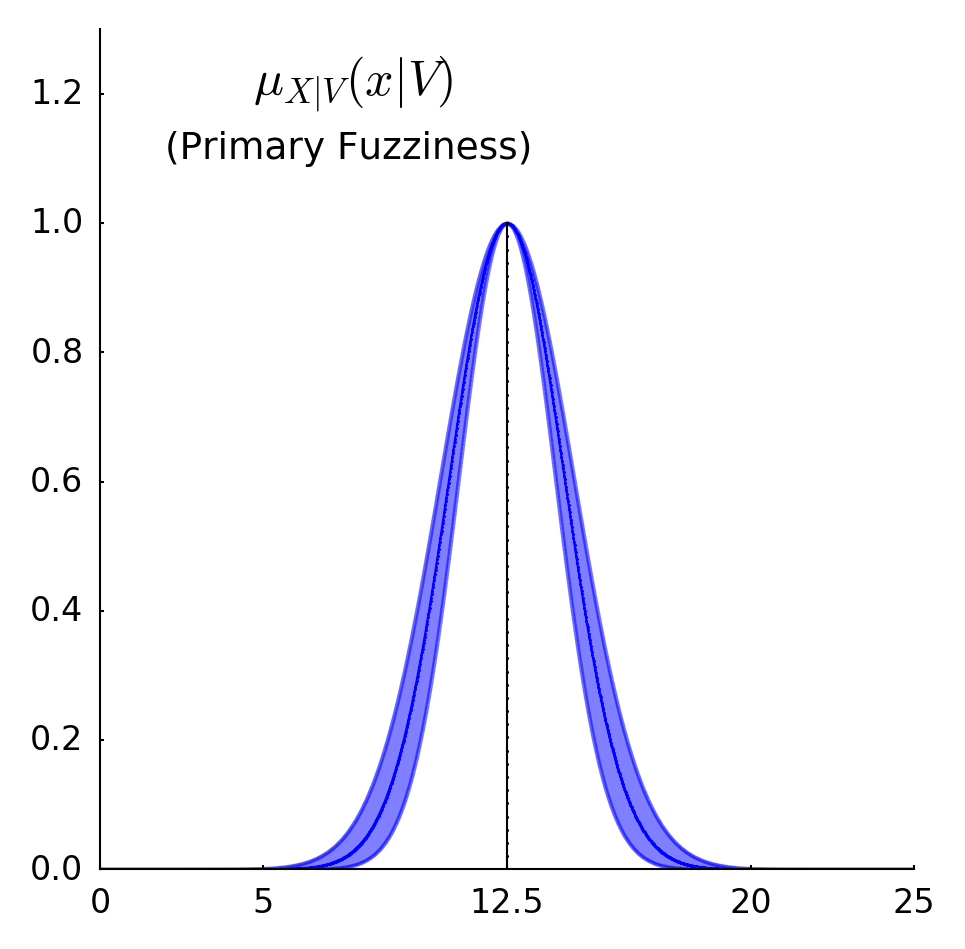}\label{primary_fuzziness}}
   \subfloat[]
    {\includegraphics[width=0.5\columnwidth]{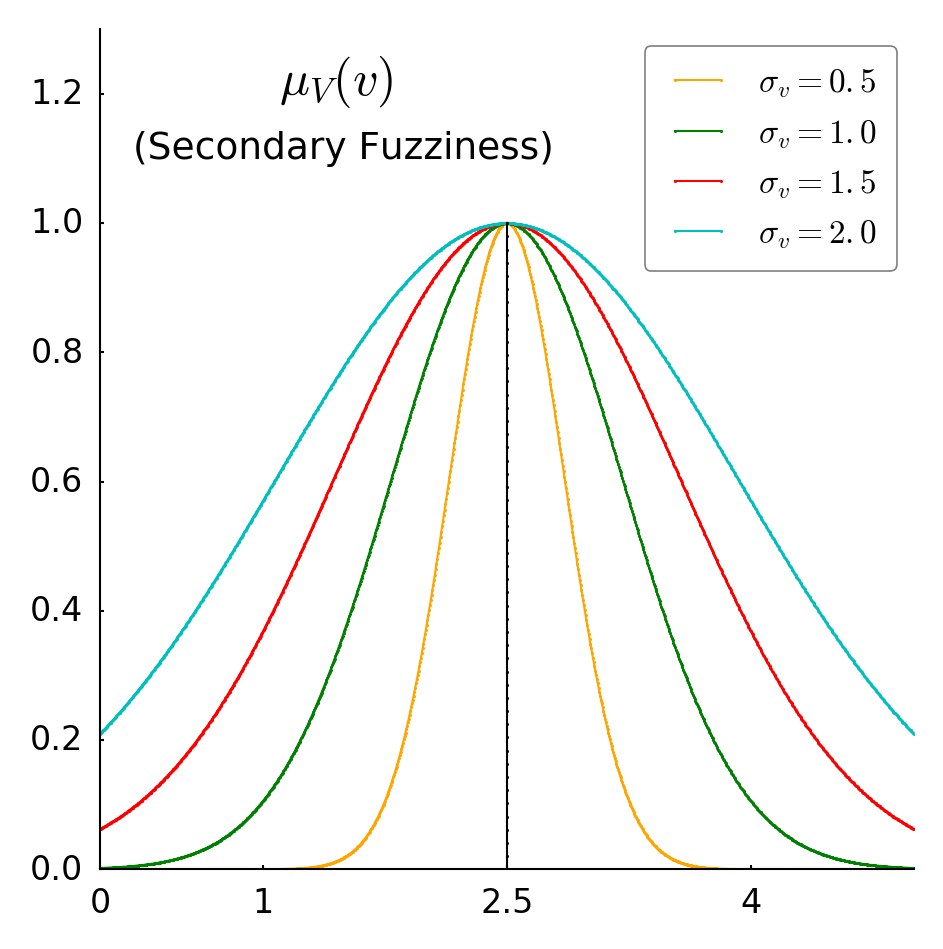}\label{secondary_fuzziness}}
   \\
   \subfloat[]
    {\includegraphics[width=0.5\columnwidth]{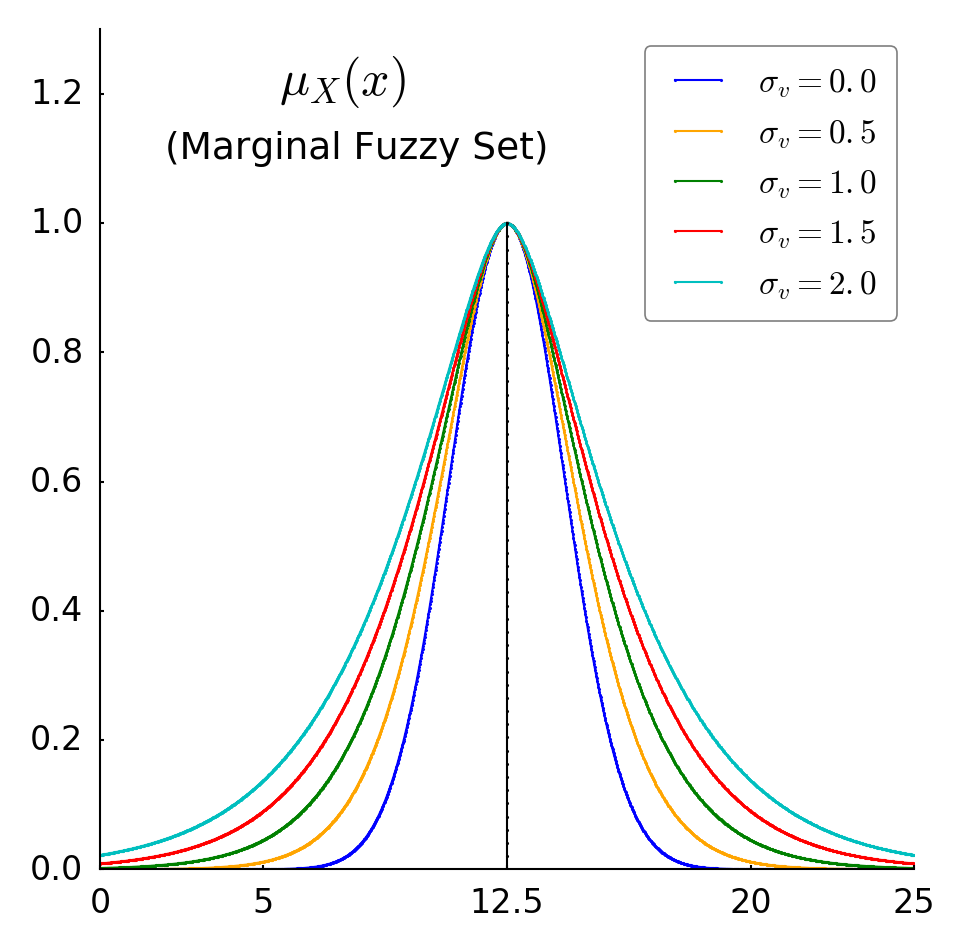}\label{marginal_fuzziness}}
  \caption{Illustration  of type-2 fuzzy set for incorporating uncertainty. (a) Primary fuzziness. (b) Secondary fuzziness with various $\sigma_v\text{s}$. (c) The final marginal fuzzy set after incorporating  uncertainty of the bandwidth with different degrees indexed by $\sigma_v$.}
\label{type2_fs_uncertainty}
\end{figure}

The above example is illustrated in Fig.\ref{type2_fs_uncertainty}. Fig.\ref{primary_fuzziness} shows the primary fuzziness when $x_0$ is estimated as 12.5 and $v_0$ is estimated as 2.5 but with uncertainty. Fig.\ref{secondary_fuzziness} shows the secondary fuzziness (uncertainty) of $v_0$ with various $\sigma_v\text{s}$. Note that we don't intend to model the uncertainty of $\sigma_v$ here. So we assume $\sigma_v$ is a given value. Fig.\ref{marginal_fuzziness} shows the marginal fuzzy set into which the uncertainty has been incorporated.

We can see from \eqref{marginal_result} and Fig.\ref{marginal_fuzziness} that the marginal fuzzy set curve is flatter when the estimated bandwidth has much uncertainty, i.e., $\sigma_v$ is large.
For a specific $\sigma_v$, the corrected bandwidth ($v_{new}$ in \eqref{marginal_result}) is almost the same as $v_0^2$ when $d(\mathbf{x}_i,\mathbf{x}_0)$ is small, and $v_{new}$ increases as $d(\mathbf{x}_i,\mathbf{x}_0)$ becomes large.
In other words, the uncertainty of the bandwidth $v_0$ is incorporated into the marginal fuzzy set $\mu_X(x)$ in such a way that membership function of points with small $d(\mathbf{x}_i,\mathbf{x}_0)$ remains almost the same shape as the one with $\sigma_v=0$ (i.e., with no uncertainty in $v_0$), and membership function of points with large $d(\mathbf{x}_i,\mathbf{x}_0)$ deviates much from the one with $\sigma_v=0$. The degree of deviation is controlled by $\sigma_v$ and $d(\mathbf{x}_i,\mathbf{x}_0)$. This behavior is very intuitive in the sense that the uncertainty of bandwidth $v_0$ is obviously reflected in the membership of $\mathbf{x}_i$ only when $\mathbf{x}_i$ is far from the center and $\mathbf{x}_i$ can be seen as a noisy datum in this case.

From the above analysis, we conclude that it's reasonable to use the marginal fuzzy set to incorporate the uncertainty of the bandwidth. But it's not easy to specify $\sigma_v$ so that the uncertainty of the bandwidth is properly represented. In next section, we will show that the choice of $\sigma_v$ depends on noise level of the data set.
\section{The Unified Framework of PCM and APCM resulted from the Uncertainty Perspective}
\label{sec-4}
In Section \ref{sec-2-2}, we propose that dataset Fig.\ref{fig1} and dataset Fig.\ref{fig6} should be dealt with differently, and that the bandwidth correction should be performed in a more flexible way. In Section \ref{sec-3-2}, we use the conditional fuzzy set formulations to implement an intuitive way of bandwidth correction. In this section, contents of previous sections are summarized. We first present the unified framework (UPCM) of PCM and APCM resulted from the uncertainty perspective. Then experiments are performed to show that the two needs in Section \ref{sec-2-2} are addressed.
\subsection{Algorithm Description}
\label{sec-4-1}
The analysis in Section \ref{sec-2-2} gives us two hints to take a new look at PCM and APCM.
First, the clustering algorithm faces a more noisy environment in Fig.\ref{fig6_ori} than in Fig.\ref{fig1_ori} because there are two close clusters in Fig.\ref{fig6_ori}. So we should have more control over the clustering process in Fig.\ref{fig6_ori}. This fact shows that each physical cluster can be seen as noise to other physical clusters.
Second, we should consider the noise existing in the data points so that we can get a reliable estimation of the membership function through the estimated uncertain bandwidth.
Coping with these two kinds of noise (uncertainty) offers us new insights into the clustering process and results in an unified framework (UPCM) of PCM and APCM.

The prototype update of each cluster is influenced by points of other clusters, in the sense that the prototype is attracted (or even dragged) by other clusters, according to \eqref{pcm_theta_update}.
Based on this observation, we introduce the concept of \emph{noise level} $\alpha$ of the data set in the update equation of prototypes:
\begin{equation}
\label{upcm_theta_update}
\boldsymbol{\theta}_j=\frac{\Sigma_{i=1}^Nu_{ij}\mathbf{x}_i}{\Sigma_{i=1}^Nu_{ij}} \quad \text{for}\;u_{ij}\geq \alpha.
\end{equation}
The $\alpha\text{-cut}$ trick is used in \cite{krishnapuram_possibilistic_1993} to compute the bandwidth with only the "good" feature point, and it's used here to update the prototype. By setting an appropriate $\alpha$, the influence of points in other clusters $\boldsymbol{\theta}_{i\neq j}$ on the $\boldsymbol{\theta}_j$ update is reduced. So we can select different $\alpha\text{s}$ for dataset Fig.\ref{fig1} and Fig.\ref{fig6}.

The uncertainty of bandwidth estimation can be attributed to the noise in data points, according to \eqref{apcm_eta_update}. Then this uncertainty causes the uncertainty of the membership value of a point through \eqref{pcm_u_update}.
In Section \ref{sec-3-2}, the intuition, that we have different confidence in the membership values of different points, is respected in the conditional fuzzy set formulation of the membership function. This conditional fuzzy set formulation \eqref{marginal_result} allows us to control the shape of the membership function through the bandwidth uncertainty parameter $\sigma_v$ in a more flexible way than simply scaling the bandwidth like \eqref{corrected_eta}.
In summary, the bandwidth $\eta_j$ is calculated with noisy points, and then the uncertainty of the membership value of a point calculated with this uncertain $\eta_j$ is modeled through the conditional fuzzy set framework. For ease of computation, we use $\boldsymbol{\theta}_j$ to replace $\boldsymbol{\mu}_j$ in \eqref{apcm_eta_update}:
\begin{equation}
\label{upcm_eta_update}
\eta_j=\frac{1}{n_j}\sum_{\mathbf{x}_i\in A_j}||\mathbf{x}_i-\boldsymbol{\theta}_j||
\end{equation}
Then update of the membership function \eqref{pcm_theta_update} is modified according to \eqref{marginal_result} as follows:
\begin{IEEEeqnarray}{ll}
\label{upcm_u_update}
\mu_{ij}=\exp\left(-\frac{d_{ij}^2}{\gamma_j}\right)
\end{IEEEeqnarray}
where $\gamma_j=\left(0.5\eta_{j}+0.5\sqrt{\eta_{j}^{2}+4\sigma_vd_{ij}}\right)^2$ and $d_{ij}=||\mathbf{x}_i-\boldsymbol{\theta}_j||$.

The above reformulation of PCM and APCM constitutes the unified framework (UPCM) for the clustering process. In UPCM, $\alpha$ and $\sigma_v$ are used together to constrain each cluster to stay in their physical clusters, and to eliminate clusters in the same dense region at the same time.
The UPCM algorithm is explicitly stated in Algorithm \ref{alg:upcm}.
\begin{algorithm}[H]
\caption{ [$\Theta$, $U$, $label$] = UPCM($X$, $m_{ini}$, $\alpha$, $\sigma_v$)}
\label{alg:upcm}
\begin{algorithmic}[1]
\Require {$X$, $m_{ini}$, $\alpha$, $\sigma_v$}
\State Run FCM.
\State Initialize $\eta_j$ via \eqref{apcm_eta_init}
\State $m=m_{ini}$
\Repeat
\State Update $U$ via \eqref{upcm_u_update}
\State Update $\Theta$ via \eqref{upcm_theta_update}
\Statex {\Comment {Possible cluster elimination}}
\For{$i \leftarrow 1 \textbf{ to } N$}
\State \textbf{Set:} $label(i)=r$ if $u_{ir}=\max_j u_{ij}$
\EndFor
\State Cluster $j$ is eliminated if $j \notin label$
\State \textbf{Set:} $m=m-p$ if  $p$ clusters are eliminated
\Statex {\Comment {Bandwidth update and possible cluster elimination}}
\State Update $\eta_j$ via \eqref{upcm_eta_update}
\State Cluster $j$ is eliminated if $\eta_j=0$ (This happens if there is only one point in Cluster $j$)
\State \textbf{Set:} $m=m-p$ if  $p$ clusters are eliminated
\Until{the change in $\theta_j$'s between two successive iterations becomes sufficiently small or the number of iterations is reached}\\
\Return {$\Theta$, $U$, $label$}
\end{algorithmic}
\end{algorithm}
\subsection{Experimental Results and Performance}
\label{sec-4-2}
In this subsection, we show the performance of UPCM on dataset \ref{fig1} and dataset \ref{fig6}. We also show the parameter-choosing flexibility endowed by UPCM.

\begin{figure}[!t]
   \centering
   \subfloat[]
    {\includegraphics[width=0.5\columnwidth]{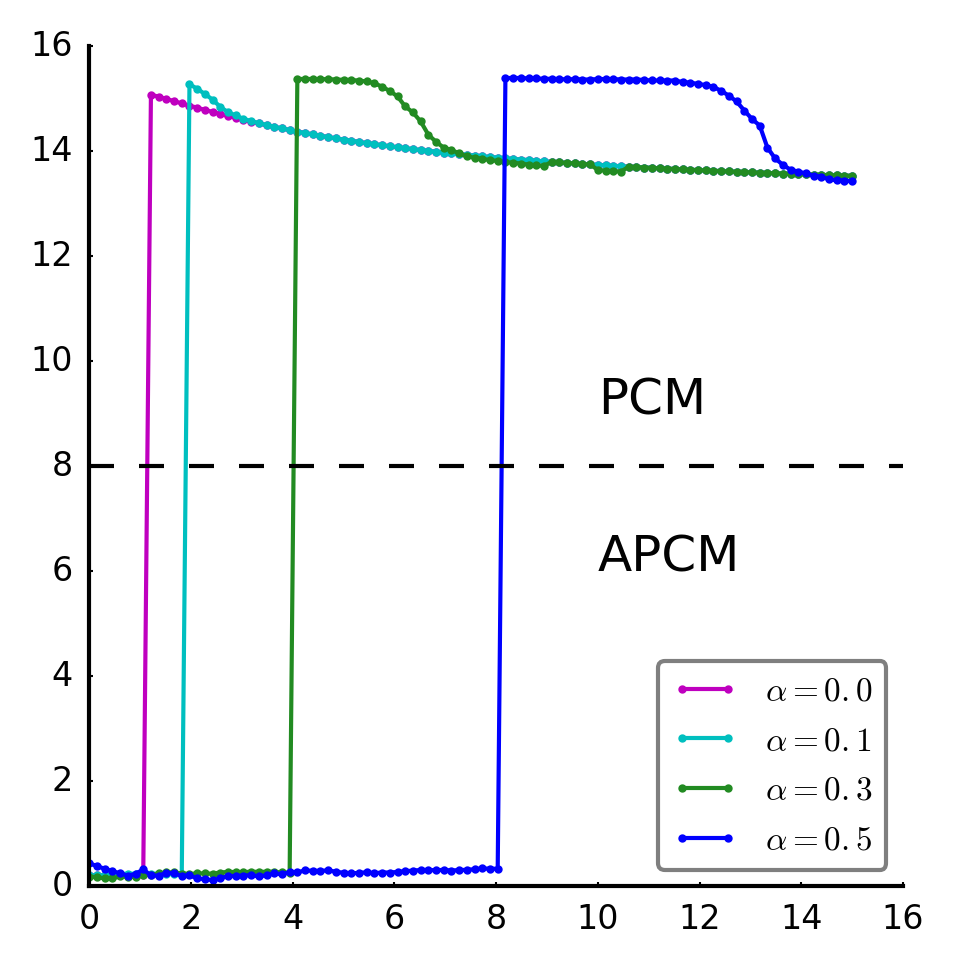}\label{fig_transition_apcm_pcm}}
   \subfloat[]
    {\includegraphics[width=0.5\columnwidth]{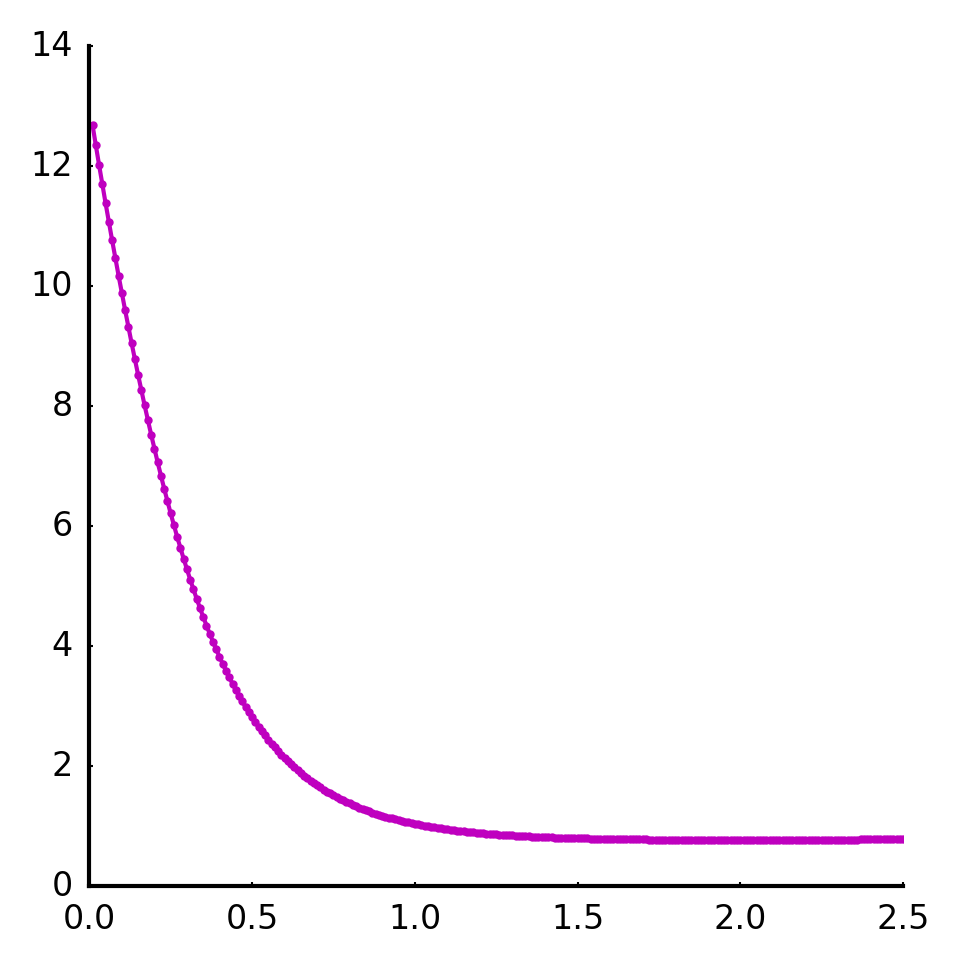}\label{fig_apcm_estimation_error}}
\caption{The center-estimation error computed via $\Sigma_j||\hat{\boldsymbol{\theta}}_j-\boldsymbol{\theta}_j^{True}||$ is used to illustrate the behavior of UPCM and APCM. Note that the estimated two cluster centers are both considered to be $\hat{\theta}_1$ when the algorithm results in only $1$ cluster to make sure it's reasonable to compute $\Sigma_j||\hat{\boldsymbol{\theta}}_j-\boldsymbol{\theta}_j^{True}||$. (a) The center-estimation error (the vertical axis) with respect to degree of uncertainty $\sigma_v$ (the horizontal axis) under various noise levels ($\alpha$). The centers are estimated by UPCM with $m_{ini}=2$ on dataset Fig.\ref{fig1}. In the APCM region, the estimated clusters are almost exactly in the two physical clusters. In the PCM region, the small cluster is "dragged" towards the large cluster. (b) The center-estimation error (the vertical axis) with respect to $\alpha$ (the horizontal axis). The centers are estimated by APCM with $m_{ini}=2$ on dataset Fig.\ref{fig1}.}
\label{estimation_error_contrast}
\end{figure}

\begin{figure}[!t]
   \centering
   \subfloat[]
    {\includegraphics[width=0.5\columnwidth]{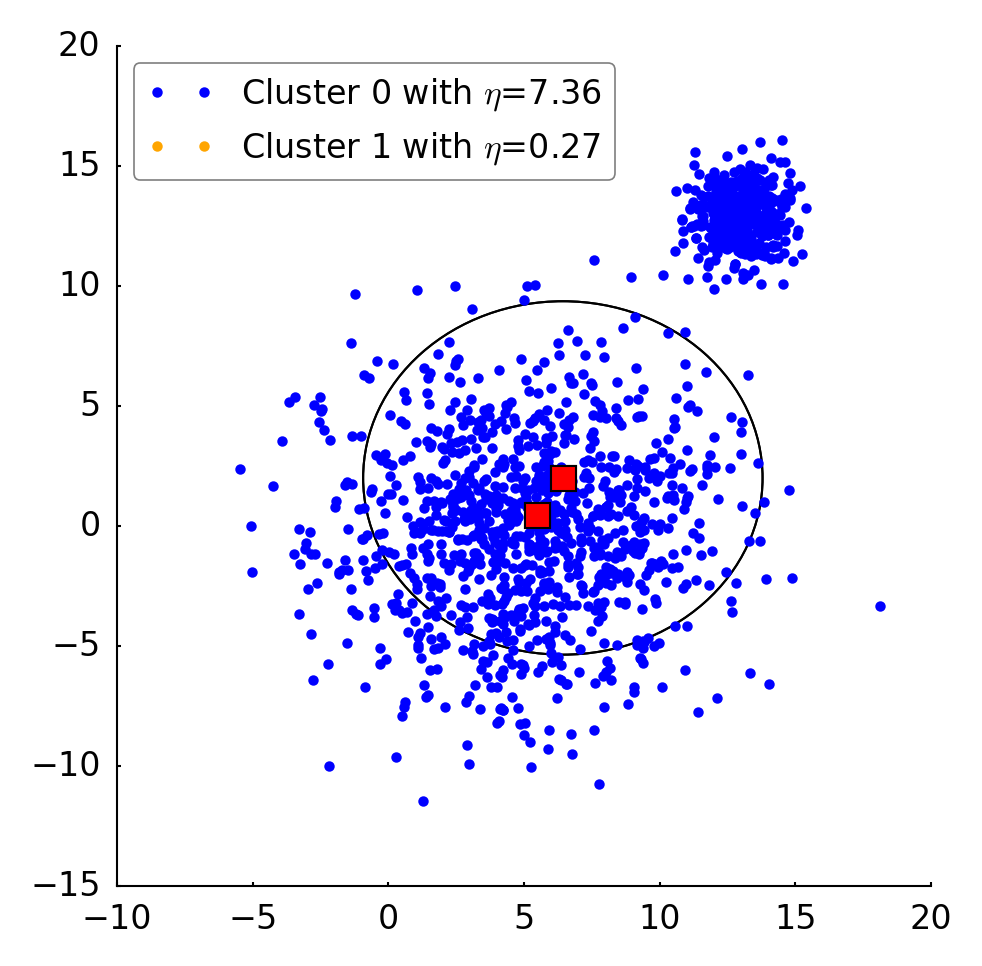}\label{fig1_notmerge}}
   \subfloat[]
    {\includegraphics[width=0.5\columnwidth]{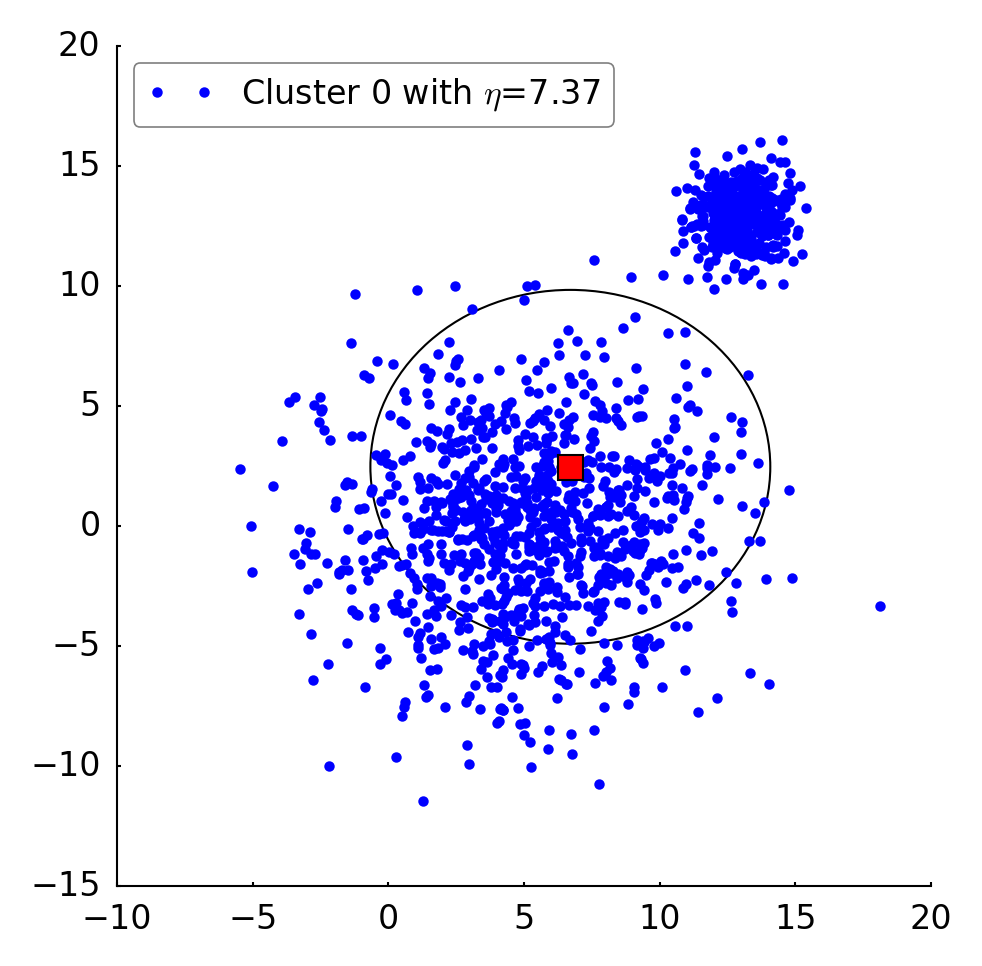}\label{fig1_merge}}
\caption{The clustering result of UPCM on dataset Fig.\ref{fig1}, in which the small cluster now has 400 points. Parameters are chosen so that UPCM operates in the PCM region corresponding to Fig.\ref{fig_transition_apcm_pcm}. (a) $m_{ini}=2$, $\alpha=4$, $\sigma_v=0$ (b) $m_{ini}=2$, $\alpha=6$, $\sigma_v=0$}
\label{fig1_merge_case_upcm}
\end{figure}

\textbf{Experiment 1:} This experiment on dataset Fig.\ref{fig1} illustrates how PCM and APCM are unified in UPCM.
Fig.\ref{estimation_error_contrast} shows the center-estimation error computed via $\Sigma_j||\hat{\boldsymbol{\theta}}_j-\boldsymbol{\theta}_j^{True}||$ with respect to parameters of UPCM and APCM.
The result of UPCM is shown in Fig.\ref{fig_transition_apcm_pcm}. The PCM region means that estimated clusters (prototypes) are both in the large cluster (Cluster 1 in Fig.\ref{fig1_ori}), and the APCM region means that estimated clusters (prototypes) are in each physical cluster respectively.
In the PCM region, the parameters given to UPCM allow the small cluster to have enough bandwidth (mobility) to move to the dense region of the whole data set (Actually, this dense region is the weighted average of points in the dataset), according to \eqref{pcm_theta_update}. At the same time, the large cluster (prototype) stays in the dense region of the large physical cluster. In other words, the small cluster is dragged towards to the large cluster. For dataset Fig.\ref{fig1} in this experiment, the two prototypes are close enough to merge into one cluster prototype when parameters (or UPCM) are in the PCM region. However, if the small physical cluster has more points, say 400, the two prototypes will merge only when we specify a large $\sigma_v$, as can be seen in Fig.\ref{fig1_merge_case_upcm}.
In the APCM region, the bandwidth (mobility) of each cluster is properly confined through $\sigma_v$ and $\alpha$, so both clusters are correctly estimated.

The result of APCM is shown in Fig.\ref{fig_apcm_estimation_error}.
We can similarly define the PCM region where two physical clusters are poorly estimated and define the APCM region where two physical clusters are well estimated.
However, the transition from PCM region to APCM region is rather smooth so that it's difficult to differentiate between these two regions.
In contrast, Fig.\ref{fig_transition_apcm_pcm} shows that the specified parameters $\alpha$ and $\sigma_v$ are either "good" or "bad" and can't have
intermediate states like "not very good". In this sense, we conclude that the main features of PCM and APCM are unified in UPCM, and that $\alpha$ and $\sigma_v$ are sufficient to control the clustering process.

\begin{figure}[htb]
\centering
\includegraphics[width=0.5\textwidth]{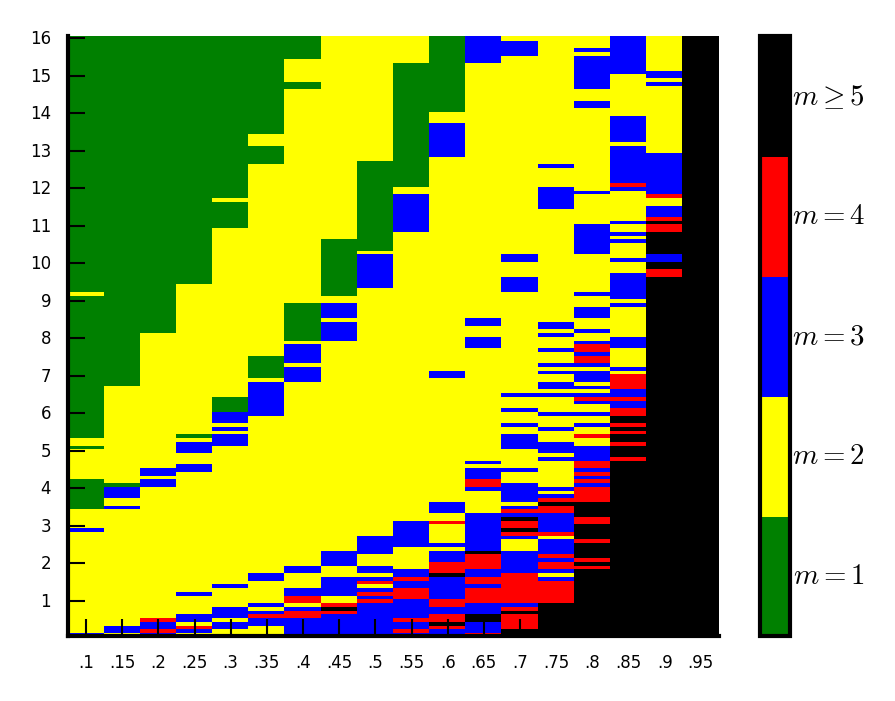}
\caption{\label{fig1_comprose}The number of clusters resulted from UPCM with $m_{ini}=10$ on dataset Fig.\ref{fig1}. The horizontal axis represents the noise level $\alpha$. The vertical axis represents the bandwidth estimation uncertainty $\sigma_v$.}
\end{figure}

\begin{figure}[htb]
\centering
\includegraphics[width=0.5\textwidth]{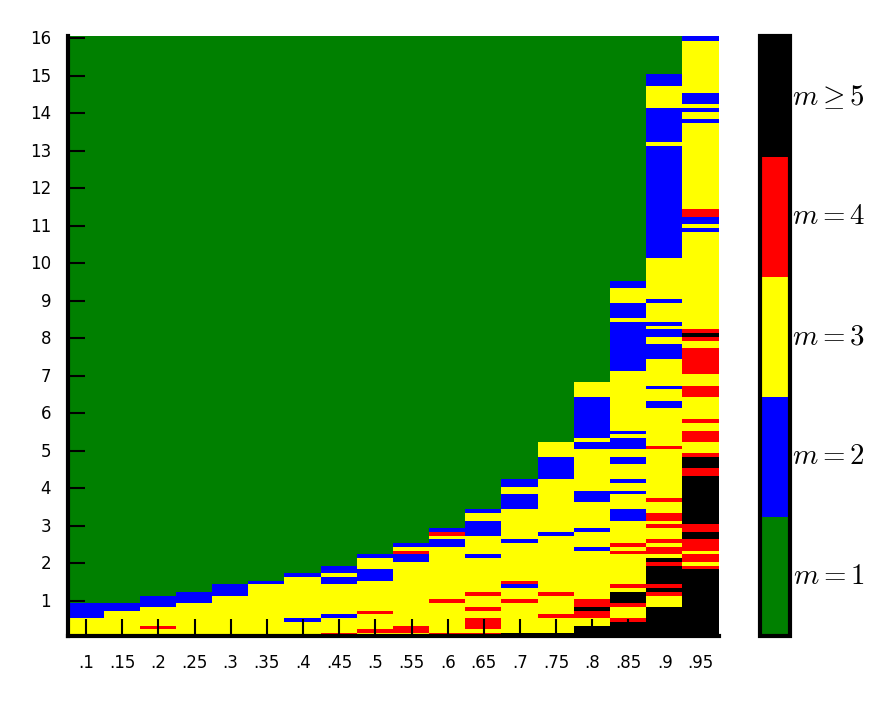}
\caption{\label{fig6_comprose}The number of clusters resulted from UPCM with $m_{ini}=10$ on dataset Fig.\ref{fig6}. The horizontal axis represents the noise level $\alpha$. The vertical axis represents the bandwidth estimation uncertainty $\sigma_v$.}
\end{figure}

\textbf{Experiment 2:} This experiment shows that the main difference between dataset Fig.\ref{fig1_ori} and Fig.\ref{fig6_ori} is the noise level.
The resulting cluster number of UPCM with $m_{ini}=10$ on dataset Fig.\ref{fig1_ori} and Fig.\ref{fig6_ori} are shown in Fig.\ref{fig1_comprose} and Fig.\ref{fig6_comprose} respectively. The results verifies that dataset Fig.\ref{fig6_ori} is more noisy than Fig.\ref{fig1_ori}.
We can see from Fig.\ref{fig6_comprose} that for the data set Fig.\ref{fig6_ori} and initialization of Fig.\ref{fig6_init}, it's better to specify a high noise level $\alpha$ so that the algorithm still estimates the correct number of clusters in a wide range of $\sigma_v$. In contrast, dataset Fig.\ref{fig1_ori} is less noisy than dataset Fig.\ref{fig6_ori} in the sense that the two clusters are not too close, so the algorithm's performance didn't rely too much on the specification of $\alpha$.

Fig.\ref{fig1_comprose} and Fig.\ref{fig6_comprose} also illustrate the interplay between $\alpha$ and $\sigma_v$, i.e., a large specification of noise level $\alpha$ indicate that fewer points are actually contributed to the adaption of prototype $\boldsymbol{\theta}_j$, so we should specify a large $\sigma_v$ to give the clusters in one physical cluster more mobility to merge. This relation between $\alpha$ and $\sigma_v$ also interprets the result of
Fig.\ref{fig_transition_apcm_pcm} where a large noise-level parameter $\alpha$ allows us to specify a wide range of $\sigma_v$ and UPCM still produces good clusters.

\section{Discussions and Further Research}
\label{sec-5}
In this paper, we analyze PCM and APCM from the perspective of uncertainty. This new perspective offers us insights into the clustering process, and also provides us greater degree of flexibility.
The proposed UPCM algorithm which comes from these insights unifies PCM and APCM in the same framework. As demonstrated on dataset Fig.\ref{fig1} in Section \ref{sec-4}, UPCM reduces to PCM when we set a small $\alpha$ or a large $\sigma_v$, and UPCM reduces to APCM when clusters are confined in their physical clusters and possible cluster elimination are ensured.

The main difference between APCM and UPCM is the way we deal with bandwidth correction.
APCM exerts strong control over the bandwidth correction process, i.e., the estimated bandwidth is directly scaled by the specified parameter \eqref{corrected_eta}, which indicates that APCM doesn't trust the estimated bandwidth. In contrast, UPCM deals with uncertainty of the bandwidth in a more fuzzy way, i.e., UPCM accepts the estimated bandwidth with a certain degree and then micro-adjusts it through $\sigma_v$. Further, the parameters of UPCM are related to uncertainty, i.e., $\alpha$ indicates noise level (uncertainty) of the dataset, and $\sigma_v$ indicates uncertainty of the estimated bandwidth. So UPCM is a PCM-based algorithm from the uncertainty point of view.

Further researches can be carried out in two directions:
\begin{enumerate}
\item The steepness of a membership function curve is controlled by the bandwidth parameter. The marginal fuzzy set \eqref{marginal_result} incorporates uncertainty of the bandwidth by making a Gaussian membership function curve flatter instead of making it steeper.
This observation leads to the conclusion that the steepness of a Gaussian membership function curve reflects uncertainty of the bandwidth. So a small cluster with small bandwidth can be interpreted as having less bandwidth uncertainty than a large cluster. In other words, the bandwidth itself can reflect the degree of uncertainty of bandwidth. Note that steepness of the membership function curve represents the degree membership values differentiate between two points. So a small bandwidth may mean that we are certain about the fact that membership values of two points are very different. This interpretation about the shape of a Gaussian  membership function can provide a new view of fuzzy clustering, and needs further investigation.
\item Based on analysis of the interplay between $\alpha$ and $\sigma_v$ in Section \ref{sec-4-2}, we can choose parameters $\alpha$ and $\sigma_v$ in the following way: specifying a small noise level $\alpha$ means that we are less uncertain about the estimated bandwidth, so we should also specify a small $\sigma_v$, and a large $\alpha$ should correspond to a large $\sigma_v$. This observation indicates that we can relate the choosing of $\sigma_v$ to noise level $\alpha$. That is, the uncertainty (fuzziness) of the bandwidth \eqref{secondary_fuzziness_v} can also be a type-2 fuzzy set with primary variable $\sigma_v$ and secondary variable $\alpha$ (see Definition \ref{conditional-fs}). Then we can cancel out  parameter $\sigma_v$ so there is only one parameter $\alpha$ for the user to choose. In this way, the clustering process is only controlled by noise level $\alpha$ of the dataset. However, modeling the relationship between $\sigma_v$ and $\alpha$ is a difficult issue and needs further research.
\end{enumerate}
\bibliographystyle{IEEEtran}
\bibliography{./ZoteroOutput,IEEEabrv}
\end{document}